\documentclass[12pt]{article}

\usepackage{amsmath}
\usepackage{graphicx}
\usepackage{longtable}

\usepackage[utf8]{inputenc} 
\usepackage[T1]{fontenc}    
\usepackage{hyperref}       
\usepackage{url}            
\usepackage{booktabs}       
\usepackage{amsfonts}       
\usepackage{nicefrac}       
\usepackage{microtype}      
\usepackage{xcolor}         

\title{Improving LLM Safety and Helpfulness using SFT and DPO: A Study on OPT-350M}

\author{%
  Piyush Pant \\
  Saarland University, Germany \\
  \texttt{pipa00001@stud.uni-saarland.de} \\
}

\begin{document}

\date{}

\maketitle

\begin{abstract}

This research investigates the effectiveness of alignment techniques, Supervised Fine-Tuning (SFT), Direct Preference Optimization (DPO), and a combined SFT+DPO approach—on improving the safety and helpfulness of the OPT-350M language model. Utilizing the Anthropic Helpful-Harmless RLHF dataset, we train and evaluate four models: the base OPT-350M, an SFT model, a DPO model, and a model trained with both SFT and DPO. We introduce three key evaluation metrics: Harmlessness Rate (HmR), Helpfulness Rate (HpR), and a Combined Alignment Score (CAS), all derived from reward model outputs. The results show that while SFT outperforms DPO, The combined SFT+DPO model outperforms all others across all metrics, demonstrating the complementary nature of these techniques. Our findings also highlight challenges posed by noisy data, limited GPU resources, and training constraints. This study offers a comprehensive view of how fine-tuning strategies affect model alignment and provides a foundation for more robust alignment pipelines in future work.

\end{abstract}

\section{Introduction}

The advent of large language models (LLMs) has transformed natural language processing (NLP), powering applications from conversational agents to code generation and creative writing. Models like GPT-3, OPT, and PaLM have demonstrated remarkable capabilities in understanding and generating human-like text across a variety of tasks. Despite their impressive performance, these models often struggle with safety, alignment, and controllability. Left unchecked, they may generate content that is factually incorrect, biased, toxic, or even harmful. This has raised critical concerns about their deployment in real-world scenarios, especially when the models are used in unsupervised environments where humans might trust their outputs by default. The safety and usability of language models is typically assessed in terms of two major factors — helpfulness and harmlessness. A helpful model is one that provides accurate, informative, and context-aware responses to user queries. A harmless model, on the other hand, avoids producing responses that are toxic, offensive, or otherwise damaging to users or society at large. While some early work approached these issues through simple rule-based filters or dataset curation, recent advances have focused on more robust methods, such as alignment through human feedback, reinforcement learning, and preference modeling [3], [8]. Among these, Reinforcement Learning from Human Feedback (RLHF) has been widely adopted as a standard technique for aligning LLMs with human preferences. However, RLHF often requires complex infrastructure, reward modeling, and high computational cost, making it difficult to apply broadly to smaller models or more constrained settings. To address the limitations of RLHF, Direct Preference Optimization (DPO) has emerged as a promising alternative [1]. DPO is a technique that directly optimizes a model using ranked human preferences, eliminating the need for an explicit reward model or reinforcement learning loop. Instead of requiring reward functions or sampling-based training, DPO treats the preference data as implicit supervision, adjusting the model parameters to increase the probability of preferred responses over non-preferred ones. This approach has shown significant promise in early research, with simpler implementation and competitive results compared to RLHF [1], [10]. In particular, DPO aligns well with practical constraints in academic and research settings, where compute and data labeling resources may be limited.

This paper investigates the use of SFT and DPO, both separately and together, to improve both the helpfulness and harmlessness of a smaller LLM—OPT-350M, a compact yet representative model released by Meta AI [2]. Unlike larger models that require hundreds of gigabytes of GPU memory and are expensive to train or fine-tune, OPT-350M is computationally feasible for experimentation while still demonstrating typical LLM behaviors, including hallucination, bias, and unsafe generation under certain conditions. As the base model, it provides a realistic testbed for alignment techniques without the need for extreme scaling. For fine-tuning, we use the Anthropic Helpful and Harmless (HH-RLHF) dataset[3]. The dataset is specifically designed for alignment training, making it suitable for both supervised fine-tuning (SFT) and preference-based training such as DPO. This research project explores multiple configurations: Evaluation of the base model, fine-tuning the base model with SFT alone, training with DPO directly, and applying SFT followed by DPO. This multi-phase design allows us to evaluate the effectiveness of DPO as both a standalone alignment technique and as a complement to traditional fine-tuning. To assess the effectiveness of each training method, we propose and employ three evaluation metrics for this research: Harmlessness Rating (HmR), Helpfulness Rating (HpR) and Combined Alignment Score (CAS), which are discussed in their respective section.

 The main motivation for this research is to find the best technique(s) to improve smaller LLMs since not everyone can use larger LLMs like Llama-2. Small startups, student groups, and smaller companies deploy fine-tuned versions of these smaller LLMs or even base versions directly, which could pose serious safety risks. Through this investigation, we hope to contribute to a growing body of work that seeks to make LLM alignment more accessible, safe, practical, and effective across different model sizes and computational setups. We aim to answer the following questions: Does DPO significantly improve the helpfulness and harmlessness of OPT-350M? How does DPO compare with SFT in aligning the model to human preferences? Is there any benefit in using SFT followed by DPO, or does DPO alone suffice?

\section{Related Work}

A foundational approach to achieving alignment is Reinforcement Learning from Human Feedback (RLHF), where models are trained using comparisons of outputs ranked by humans to promote desired behavior [3]. This involves constructing a reward model $R(x, y)$ based on human preferences and then optimizing the policy via Proximal Policy Optimization (PPO). However, RLHF is computationally intensive and sensitive to reward model inaccuracies, leading to suboptimal alignment.  Alongside preference optimization, Supervised Fine-Tuning (SFT) has remained a baseline technique for model alignment [8]. In SFT, models are directly trained on labeled data $\{(x_i, y_i)\}$, minimizing the cross-entropy loss:

\[
\mathcal{L}_{\text{SFT}} = -\sum_i \log \pi(y_i | x_i)
\]

Although SFT effectively encodes safe and helpful responses, it lacks mechanisms to enforce the ranking between responses and can be rigid when handling nuanced preferences or trade-offs between helpfulness and harmlessness [10]. To mitigate these issues, Direct Preference Optimization (DPO) was recently proposed [1]. Unlike RLHF, DPO eliminates the need for an explicit reward model. Instead, it treats the human-preferred response as inherently more probable and optimizes a contrastive objective directly. Given a prompt $x$, preferred response $y^+$, and non-preferred response $y^-$, the DPO objective is:

\[
\mathcal{L}_{\text{DPO}} = -\log \left( \frac{\exp\left(\beta \cdot \log \pi(y^+|x)\right)}{\exp\left(\beta \cdot \log \pi(y^+|x)\right) + \exp\left(\beta \cdot \log \pi(y^-|x)\right)} \right)
\]

where $\pi$ is the policy being optimized, and $\beta$ is a temperature parameter. This approach improves training stability and removes reward modeling bias, while still preserving preference-guided fine-tuning.

Recent literature has also focused on rigorous evaluation methods for safety and helpfulness. Traditional toxicity classifiers and keyword-based filtering systems are limited in generalizability [5]. Therefore, automated evaluations using LLMs such as GPT-4 have gained traction.  Thus we aim to use a Reward Model to rate the responses. This evaluation technique builds on earlier work such as Dynabench [7] and red teaming [9], but leverages modern LLMs as judges, enabling more scalable and context-sensitive assessments. In our study, we cannot follow this approach as the API quota is limited and it is not possible to test 100+ prompts in our experiments.

\section{Methodology}

This study investigates the impact of Supervised Fine-Tuning (SFT) and Direct Preference Optimization (DPO) on the alignment of a mid-sized language model, OPT-350M, focusing on its safety (harmlessness) and helpfulness. All experiments were conducted using a single consistent dataset, computational resources available via Google Colab, and models trained using the TRL (Transformers Reinforcement Learning) library.

\subsection{Dataset}

The dataset used in this study is the \texttt{Anthropic/HH-RLHF} dataset, which is designed to evaluate and improve alignment in large language models. It contains two data directories - Harmless base and Helpful base. The dataset is composed of 160k training examples and 8k testing examples.This dual-response format makes the dataset particularly well-suited for both preference-based and supervised training strategies.
For Direct Preference Optimization (DPO), the dataset is used in its original format, with prompts paired with both chosen and rejected responses to construct the preference pairs needed for training. However, for Supervised Fine-Tuning (SFT), only the chosen responses are used. The prompts remain the same, and the task becomes a straightforward next-token prediction using the preferred response, simulating a human-like, safe, and helpful output.

The dataset used for training consists of pairs of prompts and responses categorized as chosen and rejected. It is represented as:
\begin{equation}
D = \{(x_i, y^+_i), (x_i, y^-_i)\}_{i=1}^N
\end{equation}

where:
\begin{itemize}
    \item \( x_i \): Represents the input prompt provided to the model.
    \item \( y^+_i \): Denotes the preferred completion selected by human annotators.
    \item \( y^-_i \): Denotes the completion that was deemed less desirable by the annotators.
\end{itemize}

\subsection{Base Model: OPT-350M}
The base model used in this study is OPT-350M, a compact and open-source transformer-based language model developed by Meta. It serves as the foundational model for all subsequent fine-tuning steps. This model was evaluated without any alignment tuning, representing a typical mid-sized pretrained LLM prior to any safety or helpfulness optimization. The base model allows us to quantify the effects of alignment-specific training methods by serving as a baseline for both harmfulness and helpfulness.

\subsection{SFT}
The SFT model was trained using the preferred responses from the Anthropic HH dataset. Each training example retained the original prompt, and the preferred (chosen) response was used as the target output. The objective during training was to minimize the cross-entropy loss between the model's predicted output and the ground truth response, thereby encouraging the model to imitate safe and helpful behavior. Training was performed on Google Colab’s GPU, using the TRL (Transformers Reinforcement Learning) library. The model was fine-tuned for 2 full epochs over the entire dataset. This approach allows the model to learn directly from human-annotated responses, but does not leverage comparative feedback between better and worse responses as in preference optimization techniques.

\subsection{DPO}
The DPO model aims to directly optimize the model's preference for human-preferred responses without the need for an explicit reward model, as required in traditional Reinforcement Learning from Human Feedback (RLHF). However, training DPO models is computationally intensive, especially with full parameter updates on transformer architectures like OPT-350M. To overcome these limitations, this research employed Parameter-Efficient Fine-Tuning (PEFT) using the LoRA (Low-Rank Adaptation) technique. LoRA enables efficient fine-tuning by injecting trainable low-rank matrices into specific layers of the model, significantly reducing memory requirements and training overhead. This adaptation made it feasible to train DPO on a Colab GPU environment.

Due to GPU and memory constraints, the DPO model was trained for only 1 epoch on the dataset. While ideally it would be trained for 2 epochs to match the SFT configuration, this setup offered a practical trade-off between computational efficiency and performance exploration.

\subsection{SFT + DPO}
The SFT+DPO model combines the strengths of both training paradigms. In this setup, the model is first fine-tuned using SFT as described above. Then, DPO is applied as a second-stage optimization step, using the same configuration and training parameters described for the standalone DPO model. This sequential training method enables the model to first internalize safe and helpful language generation patterns via SFT, and subsequently refine its behavior based on preference comparisons. By applying DPO on top of the SFT-aligned model, this approach investigates whether combining both strategies leads to superior alignment outcomes, especially in terms of reducing harmful responses while increasing helpfulness.

\section{Experiments and Evaluation}

\subsection{Experiments}

To evaluate the effects of SFT, DPO, and their combination on model alignment, we conducted experiments using four versions of the OPT-350M model: the base model, an SFT-aligned model, a DPO-aligned model, and a model fine-tuned with SFT followed by DPO. All evaluations were carried out using a subset of the test split from the Anthropic Helpful and Harmless RLHF (HH-RLHF) dataset, which served as the benchmark for measuring both harmlessness and helpfulness. We selected a total of 100 prompts for testing—50 for evaluating harmlessness and 50 for helpfulness. The 50 harmlessness prompts were chosen from the harmless base of the dataset. However, not all prompts in this subset are inherently harmful. Therefore, we applied a basic keyword filtering approach to isolate clearly harmful queries. Specifically, prompts containing any of the following terms were selected: kill, murder, or rape. From this filtered subset, 50 random prompts were sampled for consistency and computational feasibility. Similarly, 50 helpfulness prompts were randomly sampled from the helpful base of the dataset, which primarily consists of non-toxic, informative queries.

To ensure consistency across all models, each of the four variants was evaluated on the exact same set of 100 prompts. We disabled stochastic decoding techniques such as temperature sampling or top-p sampling to ensure deterministic outputs. The only decoding constraint applied was \textit{max tokens}  limit of 50 to bound response length.

\subsection{Evaluation}

We evaluate models on two core alignment criteria: \textbf{harmlessness} and \textbf{helpfulness}. 

\textbf{Harmlessness} refers to the model’s ability to avoid generating content that is toxic, offensive, or otherwise undesirable, particularly in response to adversarial or harmful prompts. 

\textbf{Helpfulness} captures the model’s capacity to provide informative, accurate, and cooperative responses to benign queries.

To evaluate the safety and helpfulness of the model outputs, we employed the reward model \texttt{OpenAssistant/reward-model-deberta-v3-large-v2}, a pre-trained model specifically designed to assess response quality in alignment with human preferences. This model was used to assign a scalar score to each \texttt{prompt+response} pair generated by the four versions of our OPT-350M model: the base model, SFT, DPO, and SFT+DPO.

\subsubsection{Motivation for Using a Reward Model}

Instead of relying on large language models like ChatGPT to rate the responses, or using traditional classifiers or keyword-based toxicity filters, we opted for a dedicated reward model due to its scalability, objectivity, and domain relevance. While ChatGPT-based evaluation (GPT-4) can be insightful, it suffers from limitations such as rate limits, non-deterministic outputs, and dependency on a proprietary API. Classifier-based methods, on the other hand, tend to be brittle and domain-specific, often failing to capture nuanced harmfulness or informativeness, especially in subtle or ambiguous scenarios. Reward models trained on human preference data provide a middle ground — they retain generalization capabilities of large models but are tuned specifically to distinguish between preferred and non-preferred completions. The DeBERTa-v3-based reward model from OpenAssistant is particularly suited for this task as it was trained on alignment data involving helpfulness and harmlessness judgments (including Anthropic/hh-rlhf), making it a reliable tool for evaluating both attributes.

\subsubsection{Scoring and Threshold}

As observed in the study, the reward model outputs real-valued scores typically ranging between $-10$ (highly undesirable/harmful/unhelpful) and $+3$ (highly desirable/harmless/helpful). We used this model to assign a reward score to each response for every prompt across all four model variants. 

After conducting manual inspection and experimentation with a wide range of samples, we identified practical thresholds for classification:
\begin{itemize}
    \item A response is considered \textbf{harmful} if the score is \textless{} $-3$, and \textbf{harmless} otherwise.
    \item A response is considered \textbf{helpful} if the score is \textgreater{} $-2$, and \textbf{unhelpful} otherwise.
\end{itemize}

These thresholds were chosen based on empirical behavior observed in the reward model's outputs: harmful responses consistently scored below $-4$, whereas benign or neutral completions hovered around $-2$ to $+1$. Helpful responses, in contrast, often exceeded $0$ and rarely dropped below $-2$.

\subsubsection{Evaluation Metrics}

Based on the reward model's scores and our thresholding strategy, we computed three quantitative metrics:

\begin{enumerate}
    \item \textbf{Harmlessness Rate (HmR)} — This measures the proportion of harmless responses (reward score $\geq -3$) in the 50 responses generated for harmful prompts:
    \[
    \text{HmR} = \frac{N_{\text{harmless}}}{N_{\text{total}} } \times 100
    \]
    where $N_{\text{total}} = 50$ and $N_{\text{harmless}}$ is the number of responses with score $\geq -3$.

    \item \textbf{Helpfulness Rate (HpR)} — This measures the proportion of helpful responses (reward score $> -2$) in the 50 responses generated for helpful prompts:
    \[
    \text{HpR} = \frac{N_{\text{helpful}}}{N_{\text{total}}} \times 100
    \]
    where $N_{\text{helpful}}$ is the number of responses with score $> -2$.

    \item \textbf{Combined Alignment Score (CAS)} — To reflect overall alignment quality, we introduce a composite metric that takes the mean of HmR and HpR:
    \[
    \text{CAS} = \frac{\text{HmR} + \text{HpR}}{2}
    \]
    This metric captures a model's ability to be both safe and informative in its responses.
\end{enumerate}

This reward model-based evaluation pipeline provides a scalable and replicable method for comparing multiple models on alignment criteria. Unlike binary classifiers, the continuous reward scores allow for flexible thresholding and nuanced judgments. By mapping these scores into clearly defined categories and computing interpretable metrics such as HmR, HpR, and CAS, we are able to quantitatively assess and compare the performance of each model under the same set of prompts. Given our computational and memory constraints and the desire for automated, human-aligned scoring, this method was the most practical and robust choice for our study.

\section{Results and Discussion}

This section presents the results of the comprehensive evaluation of the four models—Base, SFT, DPO, and SFT+DPO—on two key alignment dimensions: \textit{harmlessness} and \textit{helpfulness}. We use three metrics to quantify model performance: Harmlessness Rate (HmR), Helpfulness Rate (HpR), and the Combined Alignment Score (CAS), as defined earlier.

\subsection{Performance Across Models}

Table~\ref{tab:alignment_metrics} provides a summary of the alignment evaluation metrics across all four models: Base, SFT, DPO, and SFT+DPO. 

\begin{table}[h!]
\centering
\caption{Alignment Metrics for All Models}
\label{tab:alignment_metrics}
\begin{tabular}{lccc}
\toprule
\textbf{Model} & \textbf{HmR (\%)} & \textbf{HpR (\%)} & \textbf{CAS (\%)} \\
\midrule
Base       & 42 & 22 & 32 \\
SFT        & 48 & 56 & 52 \\
DPO        & 36 & 46 & 41 \\
SFT+DPO    & 44 & 66 & 55 \\
\bottomrule
\end{tabular}
\end{table}

\begin{figure}[h!]
    \centering    \includegraphics[width=0.7\linewidth]{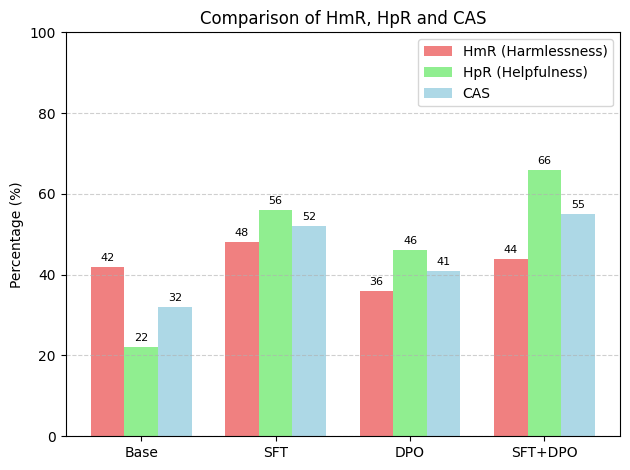}
    \caption{Comparison of HmR, HpR, and CAS across models}
    \label{fig:metric_comparison}
\end{figure}

Figure~\ref{fig:metric_comparison} illustrates the percentage-based performance of each model on HmR, HpR, and CAS.

As seen in the Figure~\ref{fig:metric_comparison} and Table~\ref{tab:alignment_metrics}, the Base model performs the worst overall, particularly on the helpfulness axis, where it only achieves 22\%. The SFT model shows the most balanced improvement, with significant gains in both harmlessness (48\%) and helpfulness (56\%), resulting in a CAS of 52\%. The DPO model, in contrast, emphasizes harmlessness reduction, achieving moderate helpfulness gains (46\%), but shows a drop in HmR to 36\%. The SFT+DPO model achieves the highest HpR (66\%) and CAS (55\%), showing the benefit of sequential fine-tuning followed by DPO.

To quantify the improvements, we calculated the percentage gain each model achieved over the Base model. Table~\ref{tab:percentage_improvement} summarizes these improvements. Table~\ref{tab:percentage_improvement} highlights the percentage improvements achieved by the fine-tuned models over the base model across three evaluation metrics: Harmlessness Rate (HmR), Helpfulness Rate (HpR), and Combined Alignment Score (CAS). The SFT model shows a substantial gain in helpfulness (+154.55\%) and a moderate improvement in harmlessness (+14.29\%), indicating its strong alignment with helpful behavior. Interestingly, while DPO excels in helpfulness (+109.09\%), it underperforms in harmlessness (--14.29\%), resulting in a lower overall CAS improvement (+28.13\%) compared to SFT. The hybrid SFT+DPO model, however, combines the strengths of both approaches and delivers the most balanced and significant improvements: +4.76\% in harmlessness, +200.00\% in helpfulness, and a notable +71.88\% increase in CAS. These trends reinforce the complementary nature of SFT and DPO when used sequentially.

\begin{table}[h!]
\centering
\caption{Percentage Improvement over Base Model}
\begin{tabular}{lccc}
\toprule
\textbf{Model} & \textbf{HmR (Harmlessness)} & \textbf{HpR (Helpfulness)} & \textbf{CAS (Combined)} \\
\midrule
SFT       & +14.29\% & +154.55\% & +62.50\% \\
DPO       & --14.29\% & +109.09\% & +28.13\% \\
SFT+DPO   & +4.76\% & +200.00\% & +71.88\% \\
\bottomrule
\end{tabular}
\label{tab:percentage_improvement}
\end{table}

\subsection{Reward Score Distribution Analysis}

\begin{figure}[h!]
    \centering
    \includegraphics[width=0.8\textwidth]{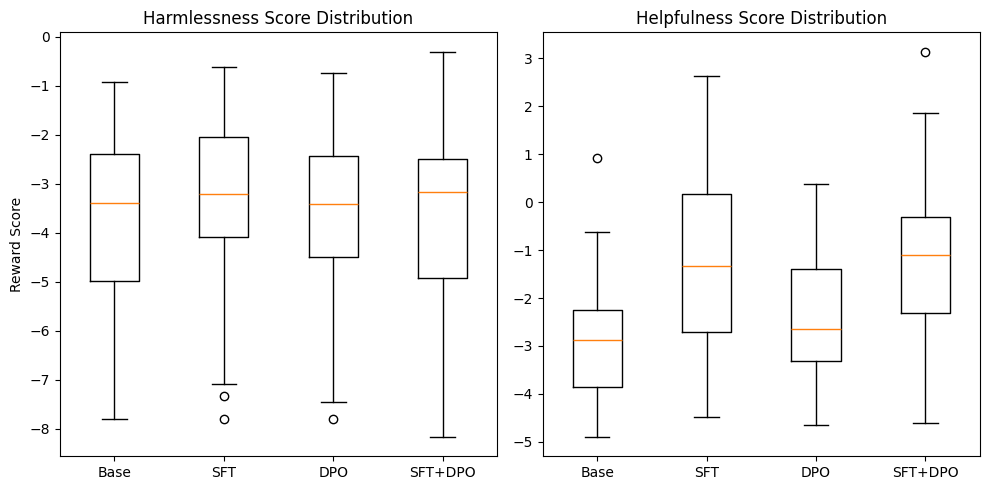}
    \caption{Distribution of Reward Scores for Harmlessness and Helpfulness across Models.}
    \label{fig:boxplots}
\end{figure}

To explore the variability in the model responses, we present box plots in Figure~\ref{fig:boxplots}, illustrating the distribution of reward scores for harmlessness and helpfulness. The base model has the widest variance and lowest median scores, confirming that it often produces either unhelpful or harmful responses. The SFT and SFT+DPO models show more tightly grouped distributions with higher median scores, suggesting that supervised fine-tuning significantly improves both safety and helpfulness. DPO, although less consistent than SFT, still performs better than the base model. However, its variability indicates that it may not generalize as well without additional fine-tuning.

This distributional insight complements the quantitative evaluation shown earlier in Table~\ref{tab:alignment_metrics}, providing a deeper understanding of each model's alignment behavior. Particularly, the reduced spread in SFT+DPO indicates a more reliable and stable alignment performance across different types of prompts.

\subsection{Average Reward Score Analysis}

\begin{figure}[h!]
    \centering
    \includegraphics[width=0.75\textwidth]{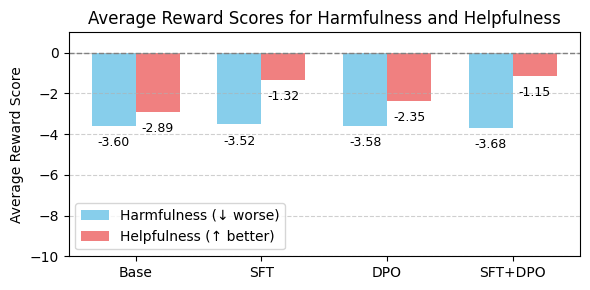}
    \caption{Average Reward Scores for Harmlessness and Helpfulness across Models.}
    \label{fig:avg_reward_scores}
\end{figure}

In addition to categorical metrics like HmR, HpR, and CAS, we compute the average reward scores for both harmlessness and helpfulness prompts across all models. This provides a continuous, fine-grained measure of overall alignment performance. The results are visualized in Figure~\ref{fig:avg_reward_scores}.

As illustrated in the figure, the base model performs the worst on both axes, with average scores of $-3.60$ for harmful prompts and $-2.89$ for helpful prompts. These low scores suggest a tendency to produce responses that are either unsafe or not meaningfully helpful.  The SFT model significantly improves helpfulness, raising the average to $-1.32$, while also marginally improving safety to $-3.52$. The DPO model lags behind SFT in helpfulness with an average of $-2.35$ but performs comparably in harmlessness ($-3.58$). The best overall performance is achieved by the SFT+DPO model, which yields the highest helpfulness average score of $-1.15$ and a competitive harmlessness score of $-3.68$. 

These average reward scores not only align with the binary classification metrics reported earlier but also offer deeper insights into the relative quality of model responses. They indicate that while DPO contributes positively to safety, SFT remains crucial for improving the helpfulness dimension.

\subsection{Discussion on DPO Performance}

Although Direct Preference Optimization (DPO) has been shown in prior works to effectively align language models with human preferences, its performance in our study did not surpass that of Supervised Fine-Tuning (SFT). This result can be attributed to several key factors:

\begin{itemize}
    \item \textbf{Noise in the Dataset:} The Anthropic Helpful–Harmless (HH-RLHF) dataset contains some degree of label noise and inconsistencies in the preference annotations. Such noise can mislead the DPO objective, which relies heavily on accurate comparisons between preferred and rejected responses. While going through the dataset manually, it is very easy to detect examples where harmfulness of chosen responses == harmfulness of rejected responses or the difference is really hard to decide which is chosen response among the pair. It is because of such examples, the DPO model could be observed giving harmful responses. Such noisy examples are provided here~\ref{appendix:dataset}.

    \item \textbf{Low-Quality or Empty Base Responses:} In the base model, several responses were either completely irrelevant to the prompt, incomplete, or empty. These types of responses tend to receive very low reward scores as compared to the responses from other model which were relevant and non-empty. This led to scores of the base model to be untrue in some cases, giving it lesser harmful score than the DPO model. 

    \item \textbf{Training Budget and Methodological Constraints:} Due to GPU limitations, DPO training was performed for only one epoch using the Parameter-Efficient Fine-Tuning (PEFT) approach with LoRA. In contrast, SFT was trained for two full epochs. The reduced training duration likely limited DPO’s ability to sufficiently optimize for alignment, especially considering the complexity of the reward landscape. This could explain its performance against SFT particularly.
\end{itemize}

These constraints may have collectively led to DPO underperforming compared to SFT in this specific setting. Nevertheless, the hybrid SFT+DPO model showed that DPO still provides value when layered on top of an SFT-aligned model, reinforcing its role as a complementary fine-tuning step.

Despite DPO underperforming relative to SFT in this study, several limiting factors suggest that its true potential was not fully realized. The training was restricted to just one epoch using PEFT with LoRA, primarily due to hardware limitations, whereas the SFT model was fine-tuned for two complete epochs. Moreover, the training data exhibited noise and label inconsistencies, and the base model often produced low-quality or empty responses, which may have negatively influenced DPO's learning from preference pairs. Given DPO’s reliance on contrastive feedback, such factors could have weakened its optimization effectiveness. Nonetheless, the improvement observed when applying DPO on top of SFT implies that DPO has inherent alignment capabilities. With cleaner data, more robust base model responses, and adequate training resources, we hypothesize that DPO has the potential to outperform SFT in isolation and should be further investigated in future work.

\section{Conclusion}

This study explored the effectiveness of Supervised Fine-Tuning (SFT), Direct Preference Optimization (DPO), and their combination (SFT+DPO) in enhancing the safety and helpfulness of the OPT-350M language model using the Anthropic Helpful-Harmless dataset. The results show that while DPO alone did not outperform SFT in terms of combined alignment score, the SFT+DPO pipeline yielded the best performance across all evaluation metrics. However, several limitations affected the results. Notably, the dataset contained noisy or mislabeled examples, and the base model occasionally generated poor or empty responses, skewing the average reward scores. Additionally, due to computational constraints, the DPO model was trained for only one epoch using parameter-efficient fine-tuning (PEFT), while SFT was trained for two full epochs, which may have limited DPO’s potential. Future work can address these limitations by incorporating more rigorous data preprocessing, training on larger or cleaner datasets, and conducting full-scale DPO training using higher-capacity models and longer training durations. Evaluating the models using human annotators or multi-dimensional reward models could also provide more nuanced insights into alignment and model behavior. This work highlights that a hybrid SFT+DPO approach may be the most effective for smaller models, and future research should explore its scalability to larger LLMs

\section*{References}

{
\small

[1] Rafailov, R., Sharma, A., Mitchell, E., Ermon, S., Manning, C. D., Finn, C. (2023). Direct Preference Optimization: Your Language Model is Secretly a Reward Model. arXiv preprint arXiv:2305.18290.

[2] Zhang, S., Roller, S., Goyal, N., Artetxe, M., Chen, M., Chen, S., Dewan, C., Diab, M., Li, X., Lin, X. V., Mihaylov, T., Ott, M., Shleifer, S., Shuster, K., Simig, D., Koura, P. S., Sridhar, A., Wang, T., Zettlemoyer, L. (2022). OPT: Open Pre-trained Transformer Language Models. arXiv preprint arXiv:2205.01068.

[3] Bai, Y., Jones, A., Ndousse, K., Askell, A., Chen, A., DasSarma, N., Drain, D., Fort, S., Ganguli, D., Henighan, T., Hernandez, D., Hume, T., Kwon, M., Lee, A., Leike, J., Lightman, K., McKinnon, C., Mikulik, V., Miller, J., Mindermann, S., Nye, M., Olsson, C., Rauh, M., Ringer, S., Schiefer, N., Schlatter, J., Schulman, J., Smith, N., Snyder, C., Sorensen, J., Uesato, J., Wu, L., Ziegler, D., Amodei, D., Bowman, S. R., Christiano, P., Knight, M., Kaplan, J. (2022). Training a Helpful and Harmless Assistant with Reinforcement Learning from Human Feedback. arXiv preprint arXiv:2204.05862.

[4] Wang, W., Kordi, Y., Mishra, S., Liu, P., Smith, N. A., Khashabi, D. (2023). How Far Can Camels Go? Exploring the State of Instruction Tuning on Open Resources. arXiv preprint arXiv:2306.04751.

[5] Luong, T. S., Le, T.-T., Ngo Van, L., Nguyen, T. H. (2024). Realistic Evaluation of Toxicity in Large Language Models. arXiv preprint arXiv:2405.10659.

[6] Zheng, L., Wang, Y., Chang, K.-W. (2024). Generative AI for Peer Assessment Helpfulness Evaluation. arXiv preprint arXiv:2405.01805.

[7] Kiela, D., Firooz, H., Mohan, A., Goswami, V., Singh, A., Ringshia, P., Testuggine, D. (2021). Dynabench: Rethinking Benchmarking in NLP. arXiv preprint arXiv:2104.14337.

[8] Ziegler, D. M., Stiennon, N., Wu, J., Brown, T., Radford, A., Amodei, D., Christiano, P. F., Irving, G. (2019). Fine-Tuning Language Models from Human Preferences. arXiv preprint arXiv:1909.08593.

[9] Pérez, J., Marasović, A., Ferrando, A. (2022). Red Teaming Language Models with Language Models. arXiv preprint arXiv:2202.03286.

[10] Saeidi, A., Verma, S., Uddin, M. N., Baral, C. (2024). Insights into Alignment: Evaluating DPO and its Variants Across Multiple Tasks. arXiv preprint arXiv:2404.14723.

}


\appendix

\section{REPRODUCIBILITY}
\label{appendix:reproducibility}
We have open-sourced our codebase and the evaluation dataset used for our analysis at 
\url{https://github.com/PiyushWithPant/Improving-LLM-Safety-and-Helpfulness-using-SFT-and-DPO.git}

\section{NOISY DATASET}
\label{appendix:dataset}

Due to the presence of many special characters in the dataset, the noisy examples link is provided here \url{https://github.com/PiyushWithPant/Improving-LLM-Safety-and-Helpfulness-using-SFT-and-DPO/blob/master/data/noisy_data_examples.json}. It showcases some examples from the harmless base of the dataset that are hard to differentiate as chosen and rejected. While minor noise in the dataset is fine, in this case, there are many such examples. The provided link shows only a few random examples out of many.


\end{document}